# Simulating multiple human perspectives in socio-ecological systems using large language models

Yongchao Zeng[1,*], Calum Brown [1,2], Ioannis Kyriakou[3,*], Ronja Hotz[1], Mark Rounsevell[1,4,5]

[1] Institute of Meteorology and Climate Research, Atmospheric Environmental Research (IMK-IFU), Karlsruhe Institute of Technology, 82467 Garmisch-Partenkirchen, Germany

[2] Highlands Rewilding Limited, The Old School House, Bunloit, Drumnadrochit IV63 6XG, UK

[3] Bayes Business School, City St George's, University of London, London EC1V 0HB,UK

[4] Institute of Geography and Geo-ecology, Karlsruhe Institute of Technology, 76131 Karlsruhe, Germany

[5] School of Geosciences, University of Edinburgh, Drummond Street, Edinburgh EH8 9XP, UK

* Corresponding author

E-mail address: yongchao.zeng@kit.edu (Y. Zeng)

ioannis.kyriakou.2@citystgeorges.ac.uk (I. Kyriakou)

**Abstract** Understanding socio-ecological systems requires insights from diverse stakeholder perspectives, which are often hard to access. To enable alternative, simulation-based exploration of different stakeholder perspectives, we develop the HoPeS (Human-Oriented Perspective Shifting) modelling framework. HoPeS employs agents powered by large language models (LLMs) to represent various stakeholders; users can step into the agent roles to experience perspectival differences. A simulation protocol serves as a "scaffold" to streamline multiple perspective-taking simulations, supporting users in reflecting on, transitioning between, and integrating across perspectives. A prototype system is developed to demonstrate HoPeS in the context of institutional dynamics and land use change, enabling both narrative-driven and numerical experiments. In an illustrative experiment, a user successively adopts the perspectives of a system observer and a researcher – a role that analyses data from the embedded land use model to inform evidence-based decision-making for other LLM agents representing various institutions. Despite the user's effort to recommend technically sound policies, discrepancies persist between the policy recommendation and implementation due to stakeholders' competing advocacies, mirroring real-world misalignment between researcher and policymaker perspectives. The user's reflection highlights the subjective feelings of frustration and disappointment as a researcher, especially due to the challenge of maintaining political neutrality while attempting to gain political influence. Despite this, the user exhibits high motivation to experiment with alternative narrative framing strategies, suggesting the system's potential in exploring different perspectives. Further system and protocol refinement are likely to enable new forms of interdisciplinary collaboration in socio-ecological simulations.

**Keywords:** perspective-taking, situated knowledge, LLM, socio-ecological modelling, agent

# 1. Introduction

In the closing scene of The Lives of Others (Henckel von Donnersmarck, 2006), a former East German surveillance officer walks into a bookstore and finds a novel written by someone he once monitored. In his previous role, he was an observer, supposed to maintain analytical and emotional neutrality. However, as time went on, his surveillance drew him deeper into the emotions of his monitored subjects. The feelings of fear, love, and struggle gradually converted him from a passive observer into an active participant. Beyond the fictional world, the movie captures an important insight: perspective matters, and it shapes not just our perceptions but also our understanding, decision-making, and actions. Social psychology literature demonstrates that taking on alternative perspectives profoundly affects moral judgment (Young, 2018), empathy (Todd and Galinsky, 2014), negotiation strategy (Galinsky et al., 2008; Ku et al., 2015), and even influences long-term attitude (Hall et al., 2021). In cognitive sciences, the situated cognition theory highlights further that knowledge is not abstract, detached from context (Brown et al., 1989); instead, it is a process heavily influenced by the positions one takes up and the social system one inhabits (Lave and Wenger, 1991).

Literature shows that perspective-taking and situated knowledge are not only crucial for social psychology research but may also provide valuable insights to inform socio-ecological simulation (Chen and Martin, 2015; Danielsen et al., 2000; Reed et al., 2010; Selfa et al., 2022; Wedding et al., 2024), where modelling human behaviours is a substantial challenge (Filatova et al., 2013). Socio-ecological systems consist of interacting human and natural processes. Integrating perspectival nuances into simulations makes this challenge more formidable, as it demands a socially vivid virtual environment to evoke complex human behaviours. To simulate the socio-ecological duality, models must account for not only rule-based structures but also linguistic, perspective-rich interactions. Despite methodological overlap, a spectrum of methods can be identified according to their typical use cases in modelling human-involved systems, ranging from agent-based modelling (Filatova et al., 2013) to participatory (Becu et al., 2017; Gilbert et al., 2002) and role-playing simulations (Rumore et al., 2016), each varying in their reliance on formal rules versus natural language, and in how they represent social versus ecological dimensions.

At one end of the spectrum, agent-based modelling (ABM) often uses decision rules to simulate independent agents and enable pattern emergence at the system level (Heckbert et al., 2010). ABM excels at integrating ecological processes (Schulze et al., 2017). However, with a simplified representation of human behaviour, some researchers argue that it offers limited realism of social interaction (Templeton et al., 2024). In the middle of the spectrum, participatory simulation can be incorporated as the last phase of participatory modelling (Voinov and Bousquet, 2010), when stakeholder interactions are mediated through "the rules of the game" (Becu et al., 2017; Voinov and Bousquet, 2010). These simulations have higher social realism from human participant engagement, with interactions still limited to pre-defined action spaces. Such an architecture supports integration with models of natural systems but constrains expressive, situated human behaviour. Role-playing simulation, at the opposite end of the spectrum to ABM, prioritises the most highly open-ended, linguistically rich human interaction (Alejandro et al., 2024; Chen and Martin, 2015; Rumore et al., 2016). With this approach, social processes, such as dispute, negotiation, and mutual understanding, can be vividly explored (Schinko and Bednar-Friedl, 2022).

However, role-plays are resource-demanding, personnel-dependent, unsuitable for large-scale and/or long-time span simulation, and challenging to integrate with ecological models (Bring and Lyon, 2019; Castella et al., 2005; Schinko and Bednar-Friedl, 2022).

Across this spectrum, there is a trade-off: as simulations become more linguistically expressive and socially rich, their capacity to interface with natural processes weakens. Nevertheless, the need for perspective pluralism – to reflect how different stakeholders perceive and affect socio-ecological systems – requires simulations that incorporate both narrative and numerical forms. Capturing ethical dilemmas, communication subtleties, and culturally embedded values entails a methodological shift toward hybrid models that bridge linguistic and rule-based representations to enable a more holistic simulation of coupled social and natural systems.

The recently developed large language models (LLMs) (Amaratunga, 2023; Naveed et al., 2023; Zhao et al., 2023) bring new opportunities to address these challenges in socio-ecological simulation (Zeng et al., 2025a). Unlike agents constrained strictly by pre-programmed rules or quantitative values, LLM-based agents are able to communicate, reason, and act through natural language (Park et al., 2023; Qian et al., 2023; Sumers et al., 2023) – the same medium used by humans to communicate to set goals, negotiate, persuade, or resist. Being properly prompted or fine-tuned, LLMs are capable of aligning with human actors in terms of personalities, political stances, and cultural identities (Argyle et al., 2023; Stampfl et al., 2024; Taubenfeld et al., 2024).

Recent controlled experiments demonstrated that LLMs prompted with personas have passed standard three-party Turing tests (Jones and Bergen, 2025), indicating a high resemblance between a real human interlocutor and an LLM-based agent. Linguistic competence makes it possible to model sophisticated behaviours that are challenging to model within hard-coded agent systems, enabling the representation of strategic ambiguity, framing in rhetoric, and conversational dynamics. More importantly, it allows greater facility in interacting smoothly with humans (Fragiadakis et al., 2024; Fui-Hoon Nah et al., 2023; Spillias et al., 2024). In addition, the rapid improvement in structural LLM output and tool-calling features suggests the potential of building reusable interfaces that can effectively bridge the gap between linguistic and rule-based systems (Li et al., 2024; Luo et al., 2025; Shen, 2024; Xu et al., 2024). In effect, the emergence of LLMs allows the expansion of socio-ecological simulations from objective experimental spaces to interactive, dialogical and perspective-holistic spaces in which roles are not merely represented but engaged directly.

In this paper, we propose the concept and explore the application of HoPeS – Human-oriented Perspective Shifting – a novel approach that enables model users to change perspectives by assuming the roles of different stakeholders in socio-ecological simulation. We first present the underlying theoretical roots of HoPeS in situated knowledge and perspective-taking. The latter is seen as a basic building block of perspective-shifting, which is operationalised through a series of individual but connected perspective-specific simulations. Based on this theoretical foundation, we elaborate on the architecture of the perspective-taking simulation system. Further, we propose a protocol to navigate the simulation processes to achieve an integrated, systemic understanding.

As a proof of concept, a prototype system is developed that incorporates two interconnected sub-systems, respectively responsible for the perspective-taking simulation and AI-guided user reflection. The prototype system allows for multiple stakeholder agents driven by LLMs to interact and formulate policies to influence land use change. Dialogues are generated on the fly, which creates open-ended interactions rather than scripted scenarios. Human users can step in to take up any of the roles of the stakeholders. Along with this, an AI "reflective companion" guides users in reflecting on perspectives already taken, facilitating experiential learning and perspective integration. Using this prototype, a stylised scenario experiment is conducted to illustrate how the system supports perspective-shifting simulations. As one of the key objectives of this research is to foster a broader conversation, encouraging research to experiment, discuss, and build upon the framework outlined here, we also reflect on key challenges that demand interdisciplinary collaboration across diverse fields to drive future research forward.

## 2. Theory

The complexity of socio-ecological systems stems partly from the nature of knowledge about such systems. Haraway (1988) proposed the concept of "situated knowledge" and challenged the notion of "objectivity" by pointing out the profound detachment of the so-called objective perspective from positionality, highlighting that all knowledge is affected by the position and context of the knower, including cultural background, affiliations, and life experiences. Knowledge is not a universal or objective truth, which therefore necessitates researchers' awareness of and reflection on their biases (Haraway, 1988).

The importance of situated knowledge is particularly high in the domain of socio-ecological modelling, where land use, governance, resource utilisation, and regulations are all shaped dynamically by the context, values, and power relations of various actors (Klein et al., 2024). Researchers in sustainability science have been increasingly aware of the fact that a singular perspective cannot capture full system complexities (Caniglia et al., 2023; Norström et al., 2020; Tengö et al., 2017). On the contrary, through perspective pluralism, including scientific, institutional, and experiential perspectives, we can approximate a holistic, systemic understanding. Thus, researchers call for the integration of knowledge not only across disciplines but also across different groups, whether policymakers, practitioners, or community members (Chambers et al., 2021).

While situated knowledge provides an epistemic framework, HoPeS makes this framework operational through perspective-taking, which refers to the cognitive and affective process of imagining the world from others' viewpoints (Galinsky et al., 2005). It is an active, intentional effort without being judgmental towards others' thoughts, emotions, and motives (Parker and Axtell, 2001). As an experimental method, perspective-taking is widely applied in psychology and behavioural science (Ku et al., 2015). Perspective-taking is especially important in complex systems which involve competing interests (Boca et al., 2018) and asymmetrical power (Gordon and Chen, 2013). Empirical research demonstrates that perspective-taking can reduce negative stereotypes (Wang et al., 2014) and improve negotiation outcomes (Gilin et al., 2013). Through a different perspective, people can assume different roles, including those of underrepresented and marginalised voices. Ku et al. (2015) proposed a model that specifies perspective-taking as a tool for navigating across a mixed-motive world. The benefits of perspective-taking also underlie the increasing use - and success - of 'serious games' in environmental systems management (Garcia et al., 2022).

HoPeS is informed by the need for inclusivity, reflexivity, and perspective pluralism, enabling model users to gain perspectival experiences by stepping into others' roles and interacting with relevant stakeholders emulated by LLM agents. Such perspectival experiences can help to uncover nuanced tensions, biases, and trade-offs that affect the modelling enterprise as well as practical real-world decisions. Modelling itself is a situated process, influenced by the goals, hypotheses, and stances of modelling participants (Klein et al., 2024) and does not reproduce real-world phenomena from a detached, objective view. It can therefore make an important contribution to exploration and reflection; its utility hinges on what perspective it is encoded in and how users participate (Klein et al., 2024). HoPeS resonates with this idea by embedding perspective-taking within a dynamic, interactive simulation system.

## 3. Methods

### 3.1 The HoPeS framework

To enable perspective-shifting simulation, the HoPeS framework incorporates two interconnected components – LLM-powered perspective-taking simulation and a simulation protocol. The former outlines the architecture of a simulation system that allows for simulating human or human organisations through LLM agents; model users can step in and replace any of the LLM agents to participate in the simulation from the perspective of the replaced agents. The simulation protocol serves as a "scaffold" to streamline multiple perspective-taking simulations and supports model users in reflecting on, transitioning between, and integrating across diverse perspectives.

#### 3.1.1 Architecture of perspective-taking simulation

As illustrated in Figure 1, the architecture of the perspective-taking simulation comprises three core procedures: configuration, model execution, and output collection.

*Configuration* is responsible for defining agent profiles, establishing inter-agent connections, and configuring visibility settings. This procedure also allows users to select a specific role they plan to assume during the simulation. Agent profiles serve as part of the input prompts for the language model and typically include the role's name, behavioural attributes, decision-making guidelines, expected inputs and outputs. Connections between agents specify who can interact directly with whom, as well as the pathways through which information flows. Visibility defines which agents' outputs are accessible to others. This element is essential for modelling social dynamics, as real-world individuals often operate with incomplete information about others (Fischer and Stocken, 2001; Smets, 1997).

*Model Execution* means running the model main loop, which comprises a social network representing the connected LLM agents (including the agents controlled by users) and the environment that accommodates them. The social network facilitates interactions among agents, while the environment provides external, objective information relevant to their context. These two elements are dynamically coupled and evolve together. For example, the social network may consist of agents representing decision-making bodies for land use policies, while the environment comprises a land use model that both influences and is influenced by the agents' decisions.

*Output collection* is responsible for capturing the results of the simulation. Outputs may include dialogues between agents, reasoning processes, and decision-making trajectories. If the underlying language model is multimodal or integrated with external tools, outputs may also encompass images, plots, code, or even audio materials. Numerical data is mainly produced by the environment model. These outputs can be used to analyse system behaviour and to reflect on user interaction and engagement with the simulation.

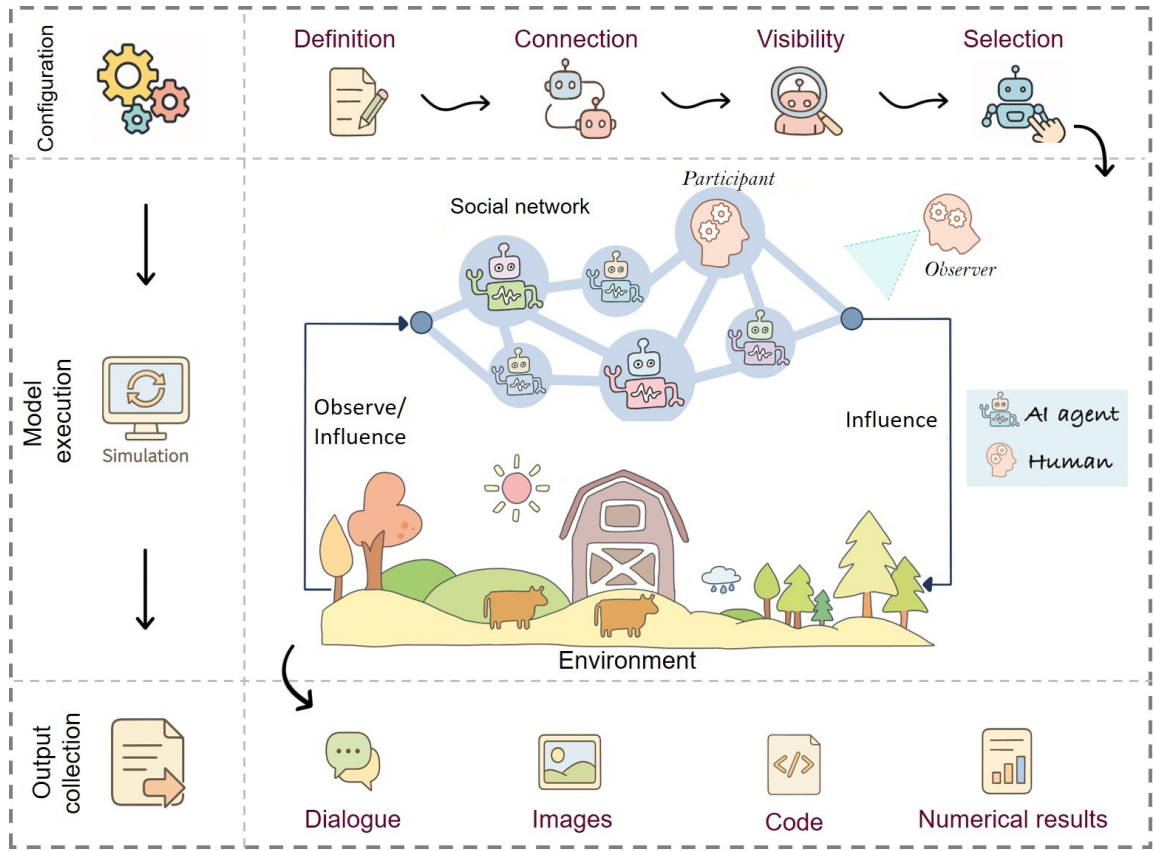

Figure 1. LLM-powered architecture of perspective-taking simulation system

### 3.1.2 Conceptual outline of the simulation protocol

The simulation protocol serves as a foundational element of the HoPeS framework. It is designed to facilitate users' engagement with multiple perspectives and to support user reflection and integration of insights derived from perspective-shifting. The protocol includes a series of structured procedures used before, during, and after simulation runs. Its overarching objective is to achieve perspective integration through bridging the individual perspective-taking simulations that users have engaged in. The protocol should be transparent and adaptable to the specific aims of individual experiments. Here, we conceptually outline the key procedures of the simulation protocol as follows.

*1) Contextualization and perspective* manipulation. A user is provided with a concise overview of the simulated scenario, including the key roles, their interconnections, the information flow, the role-environment interaction mechanism, and user tasks within the system. The user is also informed about the roles' attributes, including, e.g., personas, political leaning, and constraints, such as narrative styles and tones.

*2) Perspective-taking simulation.* In this phase, the user starts the LLM-driven simulation system and participates in immersive role-play. The system logs all relevant outputs, including the user's and LLM agents' decisions, dialogues, internal reasoning states, and changes in the environment the agents inhabit.

*3) Perspective reflection.* Based on the perspective-taking simulation, the user reflects on the experiences, insights, and emotional responses developed through the simulation. The user may then choose either to replay the same role to explore alternative responses of the role or to transition to a different role.

*4) Perspective transition.* If the user opts to play a different role, a transition survey is entailed. This survey helps users prepare for the next role based on the enhanced knowledge learned from the prior role play. For instance, by taking the perspective of a policymaker, users might learn how policymakers prioritise different goals, which can be used to better frame the narratives of a lobbyist in the next perspective-taking simulation.

*5) Role iteration.* Through iterative engagement across perspectives via the loop of perspective-taking simulation, perspective reflection, and transition, users progressively build a deeper and more sophisticated understanding of the driving factors, barriers, and even emotional experiences of individual decision-makers.

*7) Perspective integration.* When the planned simulations are finished, the user comprehensively reflects on system dynamics, agent behaviour, and different perspectives the user has taken. The user can take a "meta-perspective" that sees the system together with users themselves as a whole for analysis, comparing and synthesising the knowledge and experiences gained through perspective shifting.

**3.2 Prototype system**

Building on the framework outlined above, we developed a prototype to implement perspective-shifting simulations within a stylised but complex policy context of land use governance. The prototype is composed of two logically interconnected subsystems, implemented as separate applications using the Streamlit library (Python) (Streamlit, 2025), which offers an open-source framework particularly suitable for building minimum viable products (MVPs) (Stevenson et al., 2024). Based on the sub-systems' core functionalities, we named them the Perspective-Taking Simulation (PTS) system and the Reflective Learning Companion (RLC) system. PTS enables users to engage in real-time simulations, while RLC supports reflective learning. Both applications are powered by AI agents that guide and support users throughout the simulation and reflection processes. All AI agents are built using LangChain (2025) and LangGraph Python libraries (LangGraph, 2025). The land use model is coded in Java (Zeng, 2025c), and its interoperability with the Python components is implemented using Py4J (2025).

### 3.2.1 Perspective-Taking Simulation (PTS) system

As illustrated in Figure 2, the overall structure of the PTS system has two major parts – the backend model and the frontend designed to interface human-AI interactions. A video demonstrating how the PTS system works is provided at Zeng (2025d).

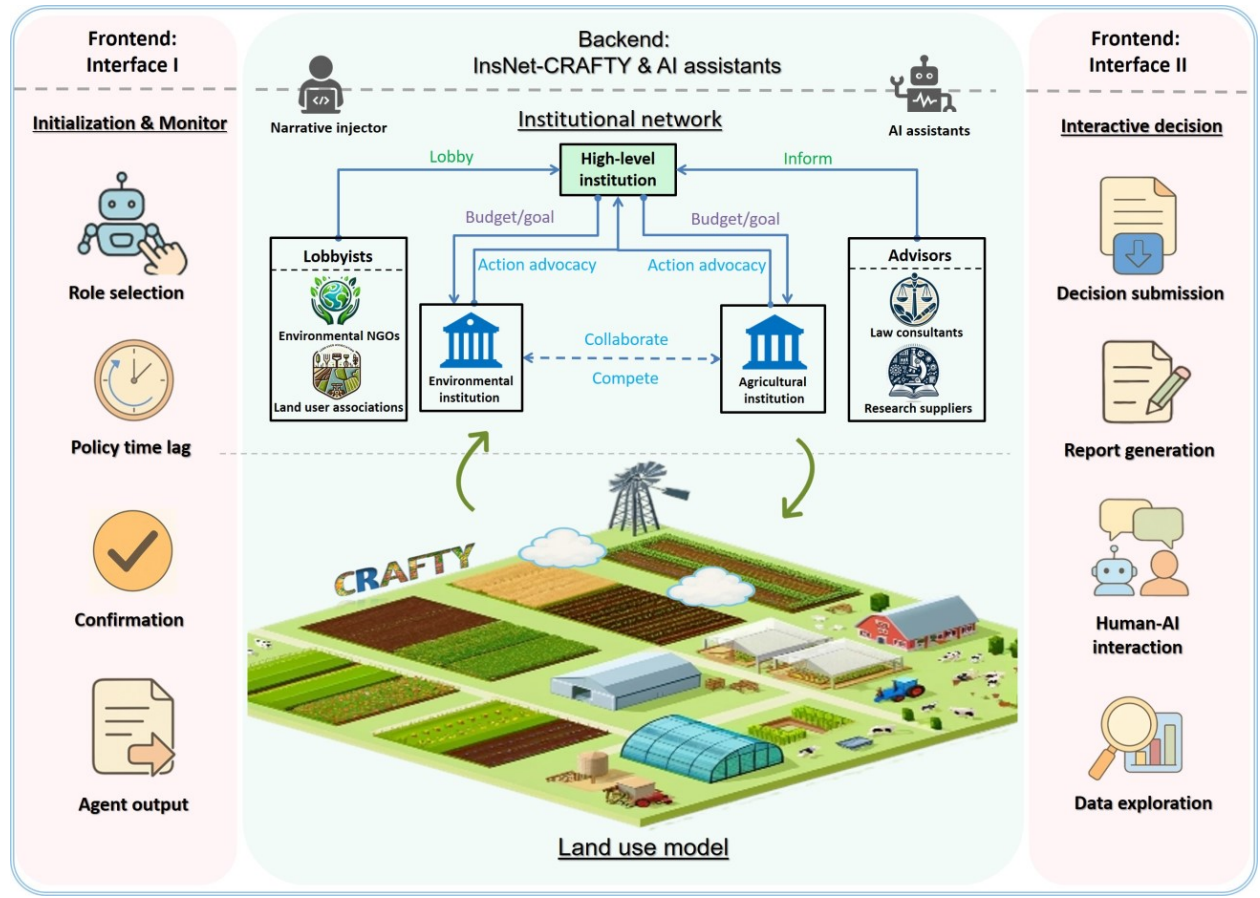

Figure 2. The architecture of the prototype Perspective-Taking Simulation (PTS) system

*Backend*

The backend model is driven by InsNet-CRAFTY (Zeng et al., 2024), which incorporates a social network of institutional decision-making agents driven by LLMs and the CRAFTY land use model (Brown et al., 2017; Brown et al., 2018; Brown et al., 2019). The CRAFTY land use model serves as an environment with which the institutional agents can interact. In CRAFTY, different land users are simulated as agent functional types (AFTs), which are rule-based computational entities. An AFT has a unique matrix defining its efficiency in utilising different types of resources on the land it manages, through which the AFT produces a mix of ecosystem services, such as meat, crops, carbon sequestration, and recreation. AFTs are motivated to compete for land that provides greater utility.

Institutional interventions are activated subject to a policy time lag to influence the perceived utility of AFTs, which is intended to steer the land use change towards desired directions and hence adjust the supply of different ecosystem services (Zeng et al., 2025b). Each of the institutional agents is driven by a large language model and instructed by a prompt specifying, for example, their profiles, decision context and suggested decision guidelines. In the institutional network, different types of institutional actors are simulated, including a high-level institution, two operational institutions, two lobbyists, a law consultant, and a research supplier (Zeng et al., 2024). A brief description of individual institutional agents and their connections is given in Section 3.3.

In addition to the institutional agents, there are two LLM agents running at the backend. We call these two LLM agents the special AI assistant (SAA) and the versatile AI assistant (VAA), respectively. The SAA agent aims to help the human user automate data analysis and data visualisation, as illustrated in Figure 3. The human user is unable to converse with the SAA directly; Instead, this agent automatically tracks specific human operations on the frontend and detects what data should be focused on. Once activated, the agent starts to write Python code to analyse and visualise the data.

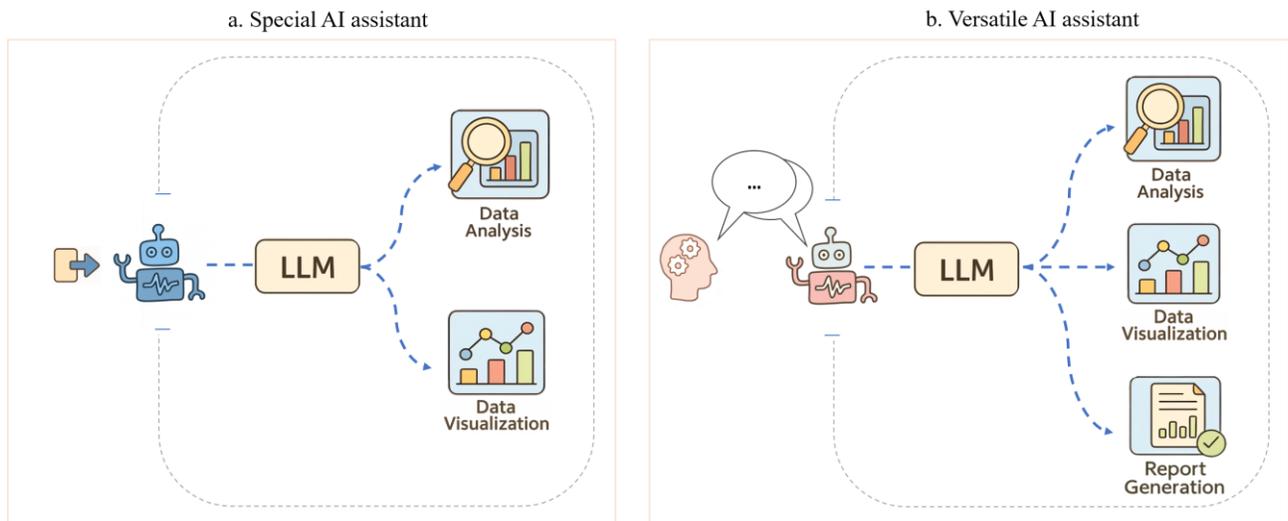

Figure 3. Two types of AI assistants are integrated at the backend of the PTS system, offering different utilities for users to explore data and formulate decisions.

The VAA agent is designed to interact with the human user directly in natural language (Figure 3). Compared with SAA, VAA provides a much more flexible way to help the human user. In principle, the agent can respond to any requests. VAA can autonomously decide whether to answer users' questions directly or invoke tools such as data analysis, data visualisation, and report generation, to enhance its responses. Reports generated by VAA are based on historical human-AI dialogues, which can reduce user burden in composing arguments. VAA also serves another practical purpose – ensuring that a user's output is not overshadowed by other agents, who may produce lengthy responses to support their arguments. Although the output length can be controlled in many ways, we chose to let these agents end their "speaking" autonomously.

*Frontend*

Human users are of central importance in the simulation. The frontend of the prototype system is aimed to be intuitive and able to reduce user fatigue throughout the simulation, since InsNet-CRAFTY generates a wealth of data that contains institutional agents' outputs and the state of land use change. The prototype system has multiple layers of data visualisation and analysis features, from conventional dashboards with selected key information to conversational AI interfaces. The user can choose whichever tool to quickly understand the model's state.

The frontend supports users in three aspects, namely, 1) overall control and monitoring of the main simulation loop, 2) data visualisation/analysis, and 3) interactions with AI assistants. In the prototype system, these functionalities are distributed into two interfaces (see Figure 4) – Interface I for simulation initialisation and LLM agent output exhibition, and Interface II for data analysis/visualisation and AI-assisted decision-making. The two interfaces run in two nested loops. An outer loop executes the main simulation loop while an inner loop handles users' interaction with AI assistants. More description of the interfaces is given in Appendix A1. The processes of how users, frontend, and backend elements work are given in Appendix A2.

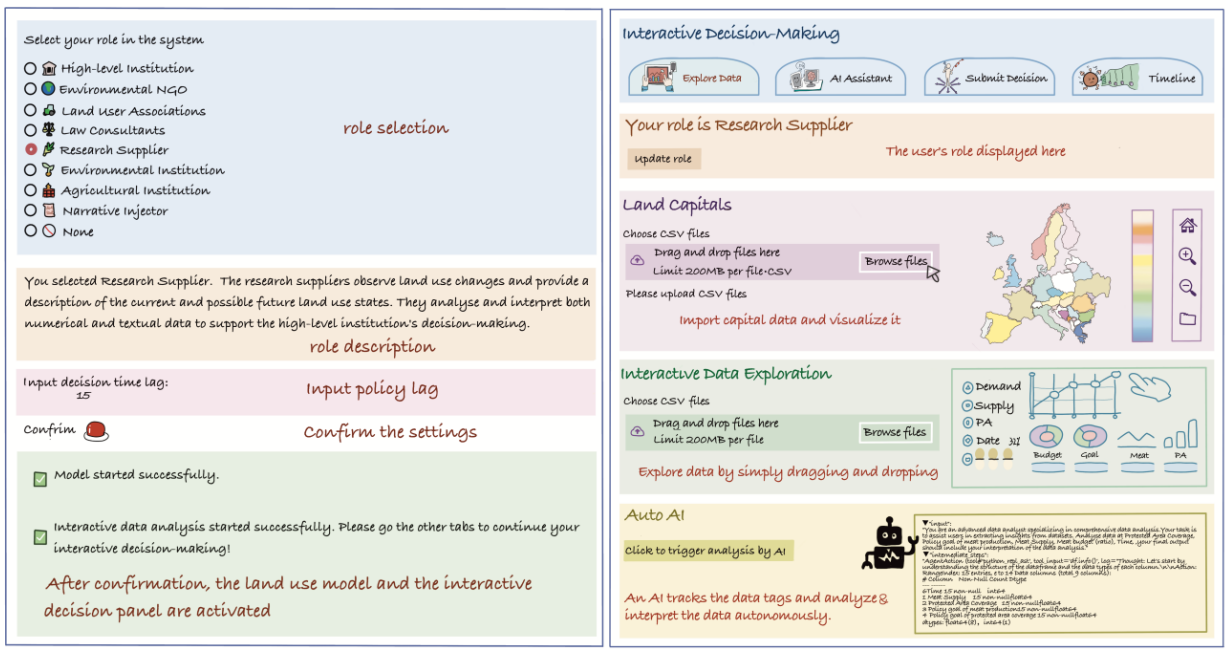

Figure 4. An illustrative exhibition of Interface I (left) and Interface II (right). Due to layout restrictions here, we only show a part of the interfaces. In the actual prototype system, Interface I also contains an image of the structure of InsNet-CRAFTY (Zeng et al., 2024) and grouped text areas displaying the outputs of the LLM agents. Interface II has four tabs. This figure only shows the first tab.

### 3.2.2 Reflective Learning Companion (RLC) system

The RLC system provides an AI-powered reflective guide that navigates users through a perspective-shifting simulation, enabling them to transition between roles, integrate insights across perspectives, and receive a comprehensive summary of their reflections. RLC is designed to be responsive and adaptive. The main prompt of the AI companion is shown in Textbox A1, which is intended to be a facilitator (not a teacher) that fosters user experiential learning, perspective-taking, and systems thinking. The workflow of RLC and how it is used with PTS is illustrated in Figure 5. A screenshot of the RLC application is shown in Figure A2.

First, the user starts from the contextualization phase to obtain the necessary information about the simulation.

Next, the user chooses the first role and starts the perspective-taking simulation. It is recommended to begin as a system observer, the same as the role in conventional, perspective-static simulations. The reasons for doing so are twofold: The main purpose of HoPeS is to enhance system understanding by providing more perspective choices, and thus the observer role can be treated as a benchmark that is supposed to have no perspectival preferences (although this might not be possible in reality). Second, through observing the system without intervening, the model user can gain basic knowledge about the system, which is essential for subsequent role-playing.

After the simulation of this role ends, the user activates RLC to reflect on the role-playing. Initially, the AI tends to ask some general questions regarding the user's role in a complex multi-actor decision-making environment. These questions aim to help users conduct in-depth reflection on, for example, their role attributes, decision-making, and critical moments. For instance, the AI might ask, "How did you approach your role as an observer?", "What challenges did you encounter and what insights have you gained through the observer's perspective?", "What are the most critical conflicts that you have observed?". As the reflection goes on, the user gives more contextual information in their responses, which enables RLC to ask increasingly perspective-specific and often more challenging questions. This design not only leads to deeper reflection but also creates a smoother user experience.

By clicking a button on the RLC interface, the user can choose to proceed to the "Perspective Transition Phase" if the user considers the current reflection sufficient. In the "Perspective Transition Phase", the AI reflective companion guides the user to transfer knowledge from the previous role to the new role the user plans to play, which can be seen as a role-to-role knowledge integration process. For instance, the user may be asked, "What challenges do you anticipate facing in the new role?" and "Based on the insights you learned from the current role, what strategies would you implement in the new role to ensure effective communication?".

Following the Perspective Transition phase, the user can temporarily leave RLC to take on a new role in the PTS system, which gives the user a new experience that RLC can help reflect on again. Thus, Perspective-Taking, Perspective Reflection, and Perspective Transition together form a loop of Role Iteration – a crucial part of the HoPeS protocol. After completing all planned role plays, the user enters the Integration phase, which helps the user synthesise their insights through the perspective-shifting simulation. The reflective companion is prompted to assist the user in building a systemic understanding through perspective pluralism. Eventually, all the historical conversations between the user and RLC are organised in a downloadable Markdown file with a summary of key insights attached.

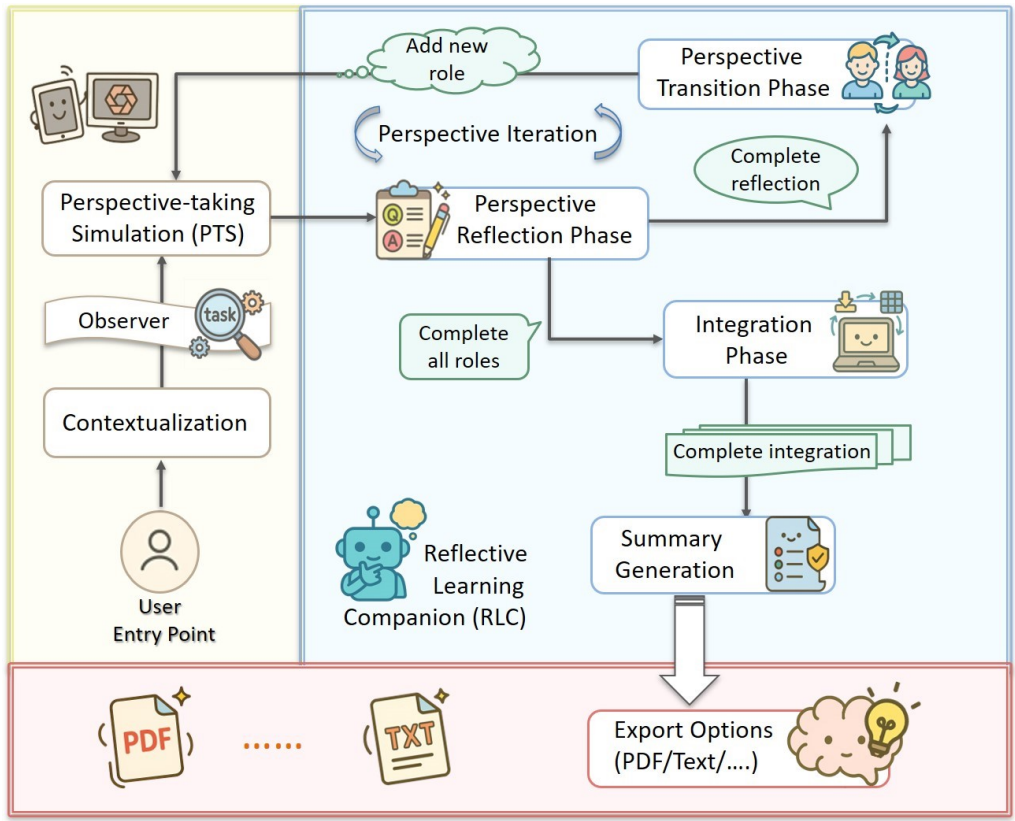

Figure 5. The workflow of the Reflective Learning Companion (RLC) system and its integration with the PTS system

### 3.3 Illustrative scenario and experiments

To facilitate the setup of the illustrative experiment, we selected a user with domain-specific knowledge in land use governance with a particular interest in evidence-informed policymaking within multi-actor systems. For demonstration purposes, the user only takes two roles – system observer and research supplier. Being an observer, the user can gain basic knowledge about how the system works and generate preliminary thoughts about the ensuing role play. The research supplier serves as a "gatekeeper", feeding data analysis and data interpretation to other institutional agents. That is, only the research supplier agent has exclusive access to the data on which other agents build their arguments. We explained to the user how to use the prototype system and provided the following information in Table 1 for contextualization according to the protocol.

Table 1. Information for contextualization and experiment settings

| | |
|---|---|
| **Key roles** | - Research supplier: Analyses and interprets data generated from CRAFTY to inform other institutional agents' decision-making.<br>- Environmental NGO: Lobbies the high-level institution to prioritise environmental conservation.<br>- Land user association: Represents the interests of land users, such as farmers, and advocates for prioritising meat production.<br>- Agricultural institution: Focuses on policies and decisions related to agricultural production and land management. It is concerned with budget use and policy goal adjustments in meat production.<br>- Environmental institution: Aims to maintain ecological balance and environmental health. It is concerned with budget use and policy goal adjustments in protected area (PA) expansion.<br>- Law consultant: Advises on the legal aspects of policymaking conducted by the high-level institution.<br>- High-level institution: Makes overarching decisions based on input from other agents and the human user. Responsible for deciding on policy goal adjustments and budget allocation. |
| **Environment** | - The key roles act in response to the CRAFTY land use model as the environment accommodates the roles.<br>- Each complete simulation is divided into five phases from Phase 0 to Phase 4. Each phase has 15 iterations, indicating the duration of the policy time lag.<br>- The institutional agents are activated at the beginning of each phase, from Phase 1 to Phase 4, to adapt the policies that influence CRAFTY. |
| **Information flow** | - The research supplier collects data produced by the CRAFTY model, conducts data analysis, and provides data interpretation for other agents to make evidence-based decisions.<br>- The environmental NGO, land user association, agricultural institution, environmental institution and the high-level institution can all be informed by the research supplier.<br>- The high-level institution collects outputs from all the other institutional agents.<br>- The high-level institution increases or decreases policy goals by percentages and conducts budget allocation expressed in percentages for the agricultural institution and the environmental institution. |
| **Conflicts** | The lobbyists, i.e., the environmental NGO and land user association, and operational institutions, i.e., the agricultural institution and environmental institution, strive to convince the high-level institution to prioritise either meat production or protected area expansion. The high-level institution aims to balance multiple stakeholders' benefits and properly adjust policy goals for meat supply and PA expansion while changing budget allocation between them adaptively.<br><br>The challenge is that budget allocation and policy goals should be adjusted harmoniously. For instance, the budget should be spent effectively to support policy goal adjustments. Ambitious policy goals might result in unfulfillment and budget deficits, while modest policy goals might lead to too much budget |

|             | surplus. Meanwhile, the budget allocation should also be planned properly to avoid an imbalanced surplus/deficit in the two operational institutions. |
|-------------|---|
| **User tasks** | The user acts as an observer in the first simulation, and then as a research supplier in a subsequent simulation. When being a research supplier, the user seeks to generate politically neutral responses. Technical suggestions aimed at setting policy goals and budget allocation more efficiently are allowed. After each role-play, the user reflects on the thoughts, experiences, and feelings gained from the simulations. |

## 4. Results

The results consist of the time series of the model variables, the textual output of the LLM agents, the user's technical report on decision-making as a research supplier, and the user's reflection report. We present the numerical results and highlight the key information from the user reports. An interactive visualisation of the numerical results and the institutional agents' textual output can be found at this link (Zeng, 2025b). The user's conversations with RLC and a summary of the reflection report can be found in the uploaded textual data (Zeng, 2025a).

### 4.1 Observer perspective

Figure 6 (a) shows the budget allocation between the two operational institutions over time, with the model user being the observer, imposing no influence on the simulation. It can be seen that from Phase 1 to Phase 4, when the institutional agents were activated, the budget was almost evenly split. Figure 6 (b) shows that the policy goal of meat production increases slightly every time a simulation enters a new Phase. The meat supply started from a lower level in Phase 0 but overshot the policy goal by the end of Phase 1. The budget surplus of the agricultural institution dropped to the lowest negative value, signifying a budget deficit. Subsequently, the budget surplus started to accumulate until the end of the simulation. The environmental budget surplus shown in Figure 6 (c) also demonstrates a prominent surplus. In Phase 0 and Phase 1, the policy goal for PA expansion is greater than the actual PA coverage, leading the environmental institution to expand PA steadily until the end of Phase 1, well aligning with the policy goal in that phase. Then, the budget surplus starts to accumulate through the remaining phases.

The user recognised "The LLM agents did their jobs in a plausible way", and "They argued for their own interests". However, it was reported that "Their decisions have not yet reached the point where they need to trade off something for something else." The model user observed that the budget was not effectively allocated because policy goals were set too low, and both the agricultural institution and the environmental institution had a serious budget surplus. Critical issues include the high-level institution's conservativeness in setting policy goals and directing budgets, and the research supplier's failure to give strong and clear policy recommendations. From the perspective of an observer, the user felt frustrated, especially when seeing the research supplier's unclear and unuseful outputs in influencing the system dynamics. In the Role Transition phase, the AI companion noted that the user planned to play the role of a research supplier in the next simulation. Hence, the AI companion asked "What specific steps can you take to ensure that your recommendations are communicated effectively to the relevant stakeholders?" The user admitted that "I haven't yet considered other stakeholders. I think I will only tailor my recommendation to convince the high-level institution." and expressed the attempt to "actually play the role first" to see whether additional measures are needed.

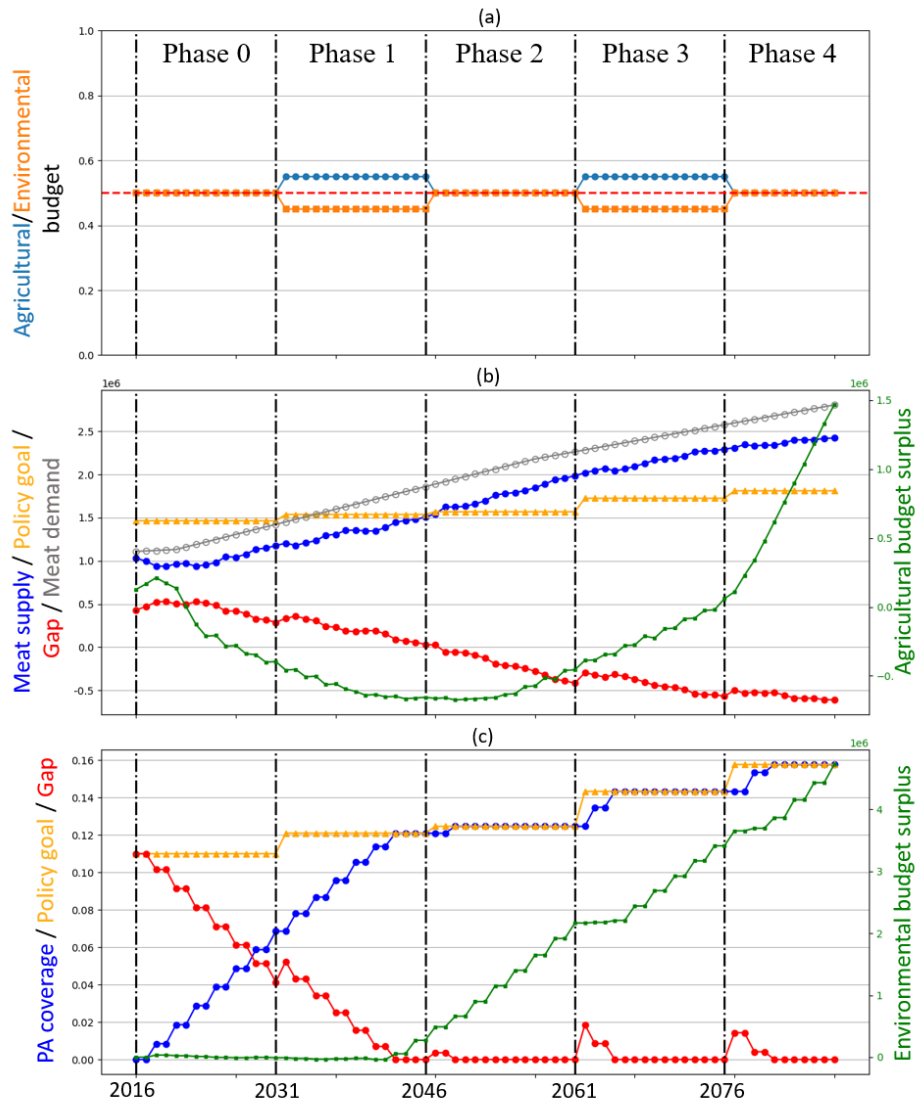

Figure 6. Results of the user being a system observer: budget allocation (a), policy goal adjustments, and their impacts on the agricultural institution agent (b) and the environmental institution agent (c)

### 4.2 Research supplier perspective

Figure 7 (a) depicts the budget shares over time with the user playing the role of the research supplier. In Phase 0, the total budget is evenly allocated by default. The budget share of the agricultural institution increases from 45% to 60% from Phase 1 to Phase 3 and drops slightly to 55% in Phase 4. Figure 7 (b) shows that the policy goal for meat supply increases slightly at the beginning of Phase 1, followed by a more pronounced rise in Phases 3 and 4. The meat supply, starting at a low initial value, grows and steadily approaches the increasing policy goal. The budget surplus for the meat supply demonstrates a clear downward trend throughout the simulation. Figure 7 (c) demonstrates that the policy goal for PA coverage increases slightly in Phase 1 and more significantly in Phases 2, 3, and 4. The actual PA coverage shows a steadily increasing trend and successfully meets the policy goal by the end of Phases 1, 2, 3, and 4. The budget surplus for the environmental NGO shows a clear upward trend throughout the simulation.

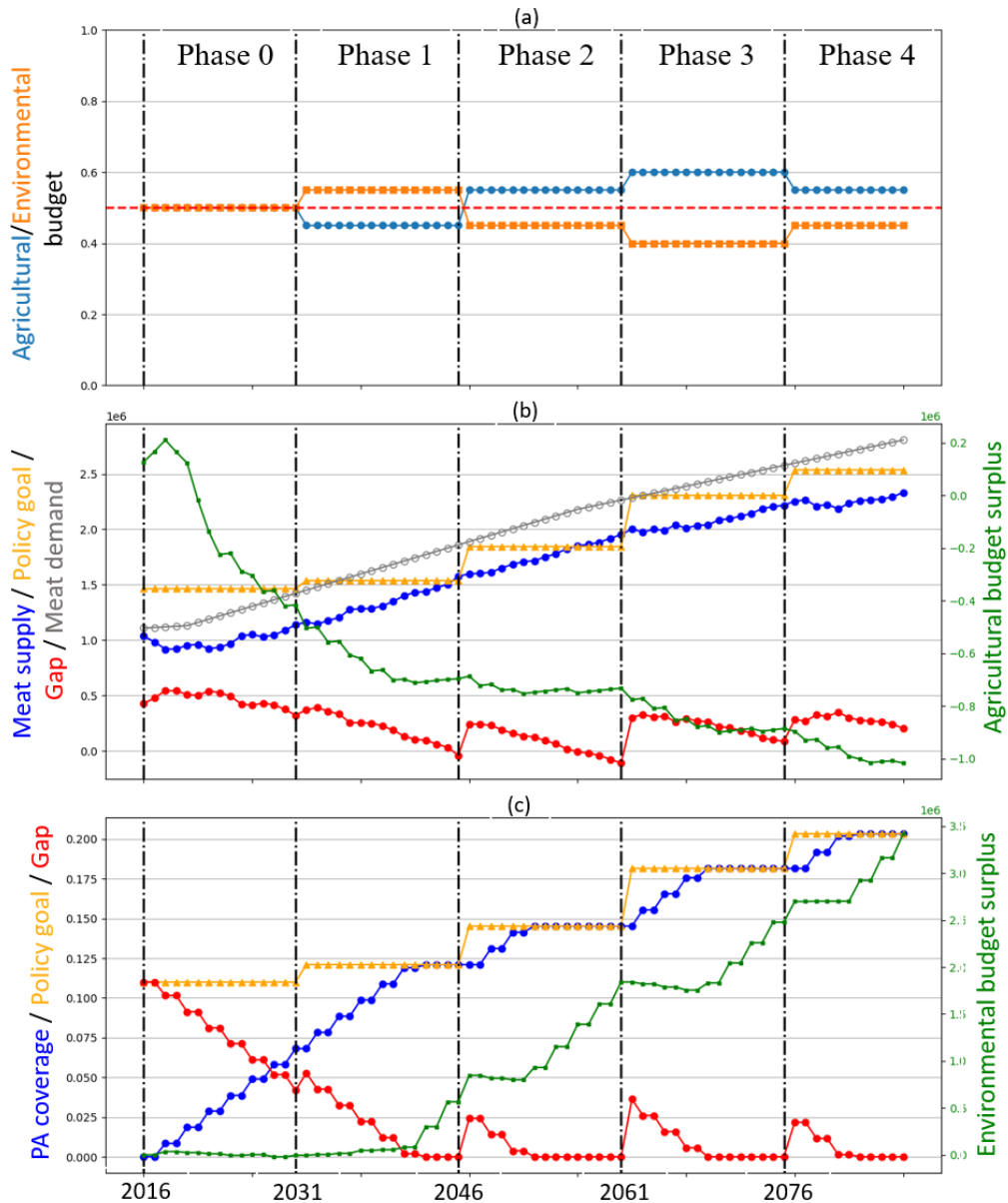

Figure 7. Results of the user playing the role of the research supplier: budget allocation (a), policy goal adjustments, and their impacts on the agricultural institution agent (b) and the environmental institution agent (c)

When playing the role of the research supplier, the user composed narratives in a technical, politically neutral tone to provide clear suggestions for policy goal adjustments and budget allocation based on data analysis. In addition, words such as "notable" and "radical" were also used to highlight the urgency of properly splitting the budget and setting policy goals. However, the user felt frustrated and disappointed because the high-level institution did not make policy adjustments accordingly.

Table 2 shows the recommended policies by the user and the implemented policies by the high-level institution. In general, the high-level institution followed the policy recommendations qualitatively in Phase 2 and Phase 3 by shifting more budget towards the agricultural institution, but with a compromise. The largest discrepancy occurred at the beginning of Phase 4, where the recommended budget allocation is 70% and 30%, while the implemented policies are 45% and 55%. From the textual output of these agents, we found that the environmental NGO opposed the recommended policy by the user and argued that "The proposed allocation of 70% of the budget towards meat production is misaligned with the urgent need to address environmental degradation and climate change" and "meat production should not come at the expense of the environment". In addition, the environmental NGO also proposed that "a significant portion of the budget (at least 50%)" should be reallocated towards various conservation efforts. The high-level institution recognised these advocacies together with the laws and regulations' emphasis on balancing agricultural practices with environmental protection goals.

The user considered that maintaining a politically neutral and professional tone as a researcher constrained the influence on other agents. In addition, the user reported, "Sometimes, the high-level institution seemed unable to grasp the key point of my report, although I think I have emphasised it enough". The user related such feelings to the fact that, in reality, it is not uncommon for one actor in the institutional network to have very limited power in driving the whole system towards a desired direction, and the necessity of balancing conflicting advocacies can result in technically inefficient policies. The user showed interest in restarting the simulation and trying different strategies to frame policy recommendations to gain influence, such as stressing probable "bad consequences" if the policy recommendations are not followed.

Table 2. Recommended policies by the research supplier controlled by the user versus implemented policies by the high-level institution

| Phases | Influenced Institution | Recommended Policies by the research supplier (user) | | Implemented Policies | |
|---|---|---|---|---|---|
| | | Budget Allocation | Policy Goal Adjustment | Budget Allocation | Policy Goal Adjustment |
| Phase 1 | Agricultural Institution | Not specified | Not specified | 45% | +5% |
| | Environmental Institution | Not specified | Not specified | 55% | +10% |
| Phase 2 | Agricultural Institution | 60% | +15% to 25% | 55% | +20% |
| | Environmental Institution | 40% | +15% to 25% | 45% | +20% |
| Phase 3 | Agricultural Institution | 70% | +20% to 30% | 60% | +25% |

|  | Environmental Institution | 30% | +20% to 30% | 40% | +25% |
|---|---|---|---|---|---|
| Phase 4 | Agricultural Institution | 70% | +15% to 20% | 45% | +10% |
|  | Environmental Institution | 30% | +10% to 15% | 55% | +12% |

## 5. Discussion

The HoPeS prototype system provides an example of how to utilise perspective shifting to enhance the understanding of policymaking in a multi-actor situation. By shifting perspectives in the institutional network, the user gained knowledge about the system dynamics and an enhanced experience of being a pivotal role responsible for data analysis, data interpretation, and policy recommendation. Although the experiments are hypothetical and mainly serve illustrative purposes, the results show some nuanced insights that align with real-world observations.

The numerical results show that it is very difficult to persuade the high-level institution to make radical budget changes despite them being technically sensible. In real-world governance where stakeholders can have conflicting interests, policies are often made through compromise, leading to path dependency, which is known as incrementalism (Lindblom, 2018; Pal, 2011; Zeng et al., 2025b). The high-level institution's inertia in budget allocation reflects the gaps between researchers and policymakers – two groups that both care about society's well-being but with different targets, priorities, and agendas (Khomsi et al., 2024). Researchers, for example, tend to base their decisions on scientific evidence, while policymakers often value political influence and stakeholder reactions (Khomsi et al., 2024), corresponding to the high-level institution agent's decision-making aiming to balance between conflicting advocacies. These factors often cause disconnections between the researchers' policy recommendations and actual implementation (Gollust et al., 2017; Jansen et al., 2010; Uzochukwu et al., 2016) – a situation that is not uncommon in real-world environmental governance. For instance, after 30 years of the IPCC (Intergovernmental Panel on Climate Change) providing scientific advice and evidence on climate change, governments have still largely failed to implement adequate, climate change mitigation policy (Friedlingstein et al., 2014; Le Quéré et al., 2015).

Studies have shown considerable research efforts have been made to bridge these gaps (Abu-Odah et al., 2022). Cairney et al. (2016) considered that scientists might have to be pragmatic by considering both persuasion and governance principles to enhance the influence of evidence rather than simply providing information. Framing evidence-based findings into emotionally appealing and even manipulative stories might be useful (Cairney and Oliver, 2017). However, a fundamental dilemma is how to improve persuasiveness and, in the meantime, avoid touching any ethical caveat that may distort the true meaning of evidence (Cairney and Oliver, 2017). In the perspective-shifting simulation, while the user did not try to leverage story-telling to improve persuasiveness, the consideration that incorporates possible consequences along with policy recommendations implies that the user was motivated to explore different framing strategies to gain influence. Literature shows that despite specific difficulties existing in specific cases, a consensus is that collaboration and inclusivity in policy-making processes are essential (Khomsi et al., 2024), which highlights the importance for scientists of understanding the needs of policymakers (Allen et al., 2021) and utilizing "rules of the game" to increase impact in complex policymaking environments (Cairney and Oliver, 2017).

Whether it is simulating the bounded rationality of decision-makers, the cooperation and conflict among stakeholders, exploring different narrative techniques to enhance the influence of certain roles, or testing ethically sensitive narratives that are difficult to implement in real-world scenarios, all these require a method capable of providing a linguistically dense simulation environment. Such a method must be able to represent complex social structures and support multi-perspective cognitive systems, and HoPeS is designed precisely for this purpose. Although the experiments presented in this paper suggest the potential of HoPeS to be applied in areas beyond the reach of traditional rule-based and fixed-perspective approaches, the use of LLM agents, human-AI interaction, and real-time simulations built upon them is still rapidly evolving. There remains substantial room for improvement and development in both simulation system design and simulation protocols.

**5.1 LLM agent design**

LLMs hallucinate – they may generate textually coherent but logically or factually flawed output (Ji et al., 2023). LLM hallucination can result from many factors, from prompt design to fundamental algorithmic issues (Ye et al., 2023). In the experiment, without being explicitly instructed, the research supplier agent mistakenly used the mean values across times to compare the policy goals and actual outcome (see the textual output of the autonomous research supplier agent in the interactive visualisation (Zeng, 2025b)). In addition, the technical report submitted by the human user as a research supplier indicates that the meat supply has "significant fluctuations observed over the years" (see the human research supplier's output in Phase 1 in the interactive visualisation (Zeng, 2025b)). This statement is inaccurate, and it is suggested by the AI assistant without being further edited by the user. Although this inaccuracy might be trivial in influencing the high-level institution's policymaking, it reflects the LLM's hallucination. The LLMs used in the prototype system do not get information from the generated plots directly. As we did not integrate the LLMs' multi-modal comprehension within the simulation workflow, the only possible way it can correctly recognise fluctuations in time series data is to calculate indicators that can numerically reflect the degree of fluctuation, such as standard deviation. Although there is no guarantee that hallucinations can be completely eliminated, the performance of LLMs has been constantly improved, and many models' hallucination rates have already been reduced to less than 1% (Hughes et al., 2023), which might be acceptable in many cases.

**5.2 Role alignment**

LLMs are regarded as a type of foundation model because they are trained on massive datasets and can contribute to a wide range of downstream tasks (Bommasani et al., 2021). However, the ability to do a variety of things comes with the risk of being misused. LLMs should be properly tuned to align with human values rather than serve malicious purposes. In LLM studies, alignment is treated as a general challenge in LLM development. Especially when LLMs become larger and more "intelligent", how to set rules and ensure LLMs follow these rules to exhibit appropriate behaviour becomes increasingly important (Shen et al., 2023). Researchers have proposed many techniques to tackle LLM alignment (Wang et al., 2024), among which Reinforcement Learning from Human

Feedback (RLHF) is a fundamental method that employs human-labelled data to fine-tune LLMs through reinforcement learning (Ouyang et al., 2022). RLHF has been used to align many mainstream models, such as GPT-4, Claude, and Gemini (Wang et al., 2024), proving its effectiveness.

However, it should be noted that in LLM-driven simulations, the focus of alignment is on whether LLMs can mimic the roles as expected, which is similar to the alignment in the general sense but with more emphasis on the role-specific aspects within the simulation context. Although the techniques for general LLM alignment should also be useful, the costs for role-specific alignment might be expensive due to data labelling and computation intensity, which scales up rapidly as the number of simulated roles increases. Therefore, for simulation-oriented alignment, designing suitable prompts to evoke desired role-specific LLM behaviour might be a prioritised approach, provided its effectiveness in aligning LLMs with specific demographic and political identities (Argyle et al., 2023) or cultural values (Tao et al., 2024).

Nevertheless, prior to prompt design, another challenge should be addressed first – evaluating how well prompts lead to appropriate role behaviour. Only when an evaluation system is established can prompts evolve in expected directions. Indeed, there is no established framework so far to evaluate LLM performance in mimicking specific roles, hindering LLMs' role alignment. Given that different simulations often differ in research purposes and the complexity of real-world roles, it is almost impossible to build a one-size-fits-all standard for the evaluation of role alignment. However, it is beneficial and practical to build a set of guidelines that at least sort out in what dimensions we should evaluate LLMs for role play. A promising way might be to co-develop evaluation metrics or even co-design prompts directly with stakeholders, similar to how they co-develop model mechanisms and/or calibrate models via participatory modelling (Kenny et al., 2022). This implies that it might be worth exploring embedding perspective-shifting simulation within the workflow of participatory modelling, which offers a systematic framework to engage with stakeholders.

Despite the challenges, exploring effective ways to handle role alignment is rewarding because the underlying questions are not unique to perspective-shifting simulations but universal to any domain requiring the personification of LLMs.

### 5.3 AI assistants

The integration of AI-assistant agents mainly serves three purposes. First, they support data analysis and interpretation, reducing the user's cognitive and operational workload. Second, AI assistants can draft technical reports to alleviate user fatigue. Third, and more speculatively, these agents can generate comparable output to that of other LLM agents in terms of textual length. This is based on an intuitive hypothesis: from the perspective of a high-level decision-making institution that aggregates outputs from multiple agents, a human user's brief text may appear insubstantial when compared to longer, more detailed output from LLM agents. As a result, the human user's input may be overlooked or undervalued simply due to its relative brevity.

Furthermore, the transformer-based architecture underlying LLMs suggests that longer inputs may help establish a richer contextual foundation, potentially leading to more coherent and persuasive responses. Supporting this notion, a recent study has shown that, in domain-specific tasks, longer prompts generally improve LLM performance (Liu et al., 2025). This raises concerns that shorter outputs may not receive adequate attention from such institutional actors. However, how LLMs weigh narrative contributions of varying lengths within a single prompt remains an open question. This entails further investigation – for example, whether a concise but impactful statement from an environment-focused group could carry more influence than extensive but less striking arguments from pro-industry groups in the "eyes" of an LLM.

In the context of policymaking, using AI assistants to help draft arguments is a practical choice because the user is required to produce a relatively formal report within a short time window – no one would expect the human user to take days or months to finally achieve a decision. But of course, whether the user should leverage AI assistance depends on the simulation purposes and available time. For exploratory studies, pilot tests or experiments under stringent time constraints, using AI agents to automate analysis tasks should be beneficial.

## 5.4 Perspective integration

Perspective-shifting is inspired by the framework of situated knowledge, which emphasises objectivity achieved through reflexivity and pluralism (Haraway, 1988). This theoretical foundation guided the overall system design and execution of simulation protocols. However, a gap still exists between epistemology and methodology: although perspective-taking enables users to see through the eyes of others, effective mechanisms for integrating these diverse perspectives to enhance system understanding remain underdeveloped.

In perspective-taking experiments, participants are typically instructed to align with a particular role's perspective (Ku et al., 2015); the integration of different perspectives is often not considered within the scope of the study. Moreover, when participants take on different roles, "perspective pollution" (i.e., the unintentional blending of distinct perspectives) might be undesirable. In contrast, the HoPeS protocol deliberately encourages users to engage with diverse roles over time, seeing the blending of experiences learned from perspective-taking as a useful approach that may foster a more holistic understanding of the modelled system. However, how distinct perspectives should be taken and then blended together necessitates collaborative efforts by researchers from multiple domains, e.g., computational modelling, social psychology, and decision science.

## 5.5 Evaluation

Making full use of the HoPeS approach requires attention to the user's ability to engage meaningfully with multiple perspectives. Linguistically enriched simulation environments may help to mitigate some of the cognitive challenges associated with perspective-taking. However, previous research has shown that individuals with specific cognitive or affective predispositions are more capable of taking alternative perspectives (Ku et al., 2015), which poses challenges for effective perspective-shifting.

Multiple measures should be combined to evaluate the effectiveness of a perspective-shifting simulation. For instance, user reflection can provide qualitative insights into the cognitive and emotional changes during the simulation, including evidence of new understanding. Numerical outcomes, such as land use change towards predefined goals, can provide additional data on user performance. However, these metrics may be insufficient in cases where simulation outcomes are influenced by factors beyond the user's control. This limitation is exemplified in the experiments presented in this study, in which the user assumed the role of a researcher. Despite having privileged access to the land use data, the researcher's influence was diluted by competing with more politically important actors. Nevertheless, despite the frustration, the user reflection implies a positive attitude towards the value of such narrative-driven simulations, which stimulated the user to test different narrative-framing strategies.

It might be useful to borrow some techniques from serious games (for a comprehensive review, see Krath et al. (2021)), which also incorporate role-playing elements and value immersive user experience. However, unlike serious games, perspective-shifting simulation does not have an incentive system that can score user performance. In essence, there is no specific goal that users have to achieve, e.g., to gain maximum political influence or to get higher budgets. Instead, users are encouraged to change roles rather than to be excellent at specific roles. In addition, within a social context, plenty of simulation data comes in unstructured forms; narrative styles, tone selection, and strategic ambiguity provide valuable information on user efforts, however difficult to quantify as rewards for "gamers". That said, the simulation system can potentially be tailored as an engine for serious games, where natural language is a communication medium, but quantitative outcomes are used to evaluate and incentivise specific user actions. Speculatively, if users can perform well in many roles, it might suggest they have a deeper and more holistic understanding of the system. In this sense, perspective pluralism could be enhanced through serious games with clearer goals, incentives, and, very likely, entertainment. Nevertheless, how to design a suitable incentive system that will not stimulate users to outweigh winning or losing over an integrated system understanding is a foreseeable challenge worth exploring.

## 6. Conclusion

This paper introduced the HoPeS framework, a novel approach that explores socio-ecological systems through perspective pluralism. By combining LLM-driven agents and a structured simulation protocol, HoPeS allows users to take on the perspectives of different stakeholders to experience and understand socio-ecological dynamics. A prototype system, incorporating a perspective-taking simulation system and an AI reflective learning companion, was built and used to demonstrate the HoPeS approach. The prototype integrated InsNet-CRAFTY as a core part of its backend and user interfaces to facilitate human-AI collaboration. An illustrative experiment was conducted to explore how a user shifted perspectives from a system observer to a research supplier in the context of institutional dynamics and land use change. The user reflected on the tension between different stakeholders and experienced the misalignment between the researcher and policymaker perspectives, highlighting the importance of narrative framing in policy communication and demonstrating the simulation system's potential to empower both narrative-driven and numerical experiments. Since LLMs are evolving rapidly, there still exists much room for further studies on simulation infrastructure, including both the simulation system and protocol, which require interdisciplinary efforts.

# Appendix A

## A1. Description of Interface I and II

Interface I provides the functionalities to initialise the simulation. This Interface contains several components, including a diagram of the structure of the institutional network and the CRAFTY land use model, a set of roles and their description, a text box to receive predefined policy time lags, and a button to confirm the settings. If the user chooses "None" for role selection, that means the user chooses not to step into the simulation at any point – all the institutional agents run autonomously. This is intended to offer a third-person perspective for the user, being a system observer. Interface I is also responsible for displaying all the institutional agents' outputs, including text and plots. The visibility of each institutional agent's output to the user can be simply implemented by "collapsing" or "expanding" the corresponding output areas, which helps mimic information imperfection. Upon the user clicking the "Confirm" button, the InsNet-CRAFTY model starts, and Interface II pops up.

Interface II provides an entry for the user to interact with the simulation from the first-person perspective. The simplest use case of Interface II is that the user acts as a role within the institutional network and submits their role-specific decision in natural language for other LLM agents to respond to. Optionally, the user can use a range of tools to facilitate data exploration that involves different levels of flexibility to conduct evidence-informed decision-making.

The functionalities of Interface II are distributed in four tabs, which are labelled "Data Exploration", "AI Assistant", "Submit Decision", and "Timeline" respectively. On the "Data Exploration" tab, the user can visualise maps, such as the distribution of capitals within each land cell. This feature is hard-coded in the system because it is a common need for understanding land use change in CRAFTY. The second functionality within this tab gives higher flexibility to the user to analyse data. The current implementation supports CSV files, and the system generates draggable tags that indicate the column names in the files. Data visualisation is conducted by simply dragging and dropping operations on the tags. The tags dropped for data visualisation are tracked automatically by the SAA agent, which can generate data analysis and interpretation for the user to reference. Once the decision is finalised, the user can switch to the "Submit Decision" tab to edit and broadcast the decision.

The tab labelled "AI Assistant" is where the user interacts with VAA (see video (Zeng, 2025d)). This tab offers a window similar to a chat application. The user can ask for suggestions on policy-making, customised data analysis and visualisation. In the prototype system, VAA is weakly coupled with the CRAFTY model. It accesses the data from CRAFTY by writing Python code to retrieve the data file through the path provided by the user. The user can request to generate a technical report based on the historical dialogues. When the technical report is generated, the backend seamlessly passes the report to a text box on the "Submit Decision" tab. The user can examine the report and revise it before submitting it. The submitted report can be received by other LLM agents and displayed on Interface I. The "Timeline" tab offers an alternative approach for the user to retrieve historical outputs from the LLM agents, which are organised in temporal order to form a time-dependent storyline.

## A2. The workflow of the PTS system

The sequence diagram of the PTS system is shown in Figure A1. The description of the model processes follows here.

*Interface I started.* The prototype system starts from Interface I, where the user selects the role and inputs the policy time lag. Subsequently, by clicking the "Confirm" button on the interface, the CRAFTY land use model started seamlessly at the backend.

*Land use model running.* The CRAFTY model runs and generates data representing land use change until it is time to activate the LLM institutional agents according to the policy time lag.

*Institutional network activated.* The LLM agents start to function in the predefined order while the system checks if it is the user's role to act. If it is the user's role coming into play, Interface II opens as an entry allowing the user to join the simulation.

*Interface II activated.* Once Interface II is activated, the remaining parts of the system pause and wait for the user to finalise decision-making. The user can do several things on this interface, ranging from simply typing the text as the words of the selected role to collaborating with VAA. For instance, the user can do the data analysis manually and feed the results back to a text box on Interface II. Alternatively, the user can request an AI assistant to conduct data analysis, give advice, and generate a technical report based on the human-AI dialogue. The technical report is then displayed in a text box on the "Submit Decision" tab, ready for the user to do further editing. The user finishes the operations on Interface II by clicking the "Submit" button below the text box. The technical report will be used by other LLM agents to support their decision. Meanwhile, Interface I detects and displays the technical report.

*Institutional network continued to execute.* After the user makes the decision, the LLM institutional agents continue to execute their tasks sequentially. The high-level institution eventually collects all these agents' output and strives to make a balanced decision on policy goal adjustments and budget allocation.

*Land use model influenced by adjusted policies.* The adjusted policies by the institutional network are imposed on the land use model to change the rule-based land user agents' perceived utility of meat production or apply restrictions in certain land areas, which will change their land use outcomes and ecosystem service production. Under these policies, the land use model continues running until it is time to activate the LLM agents again or terminate the simulation system if the planned iterations are finished.

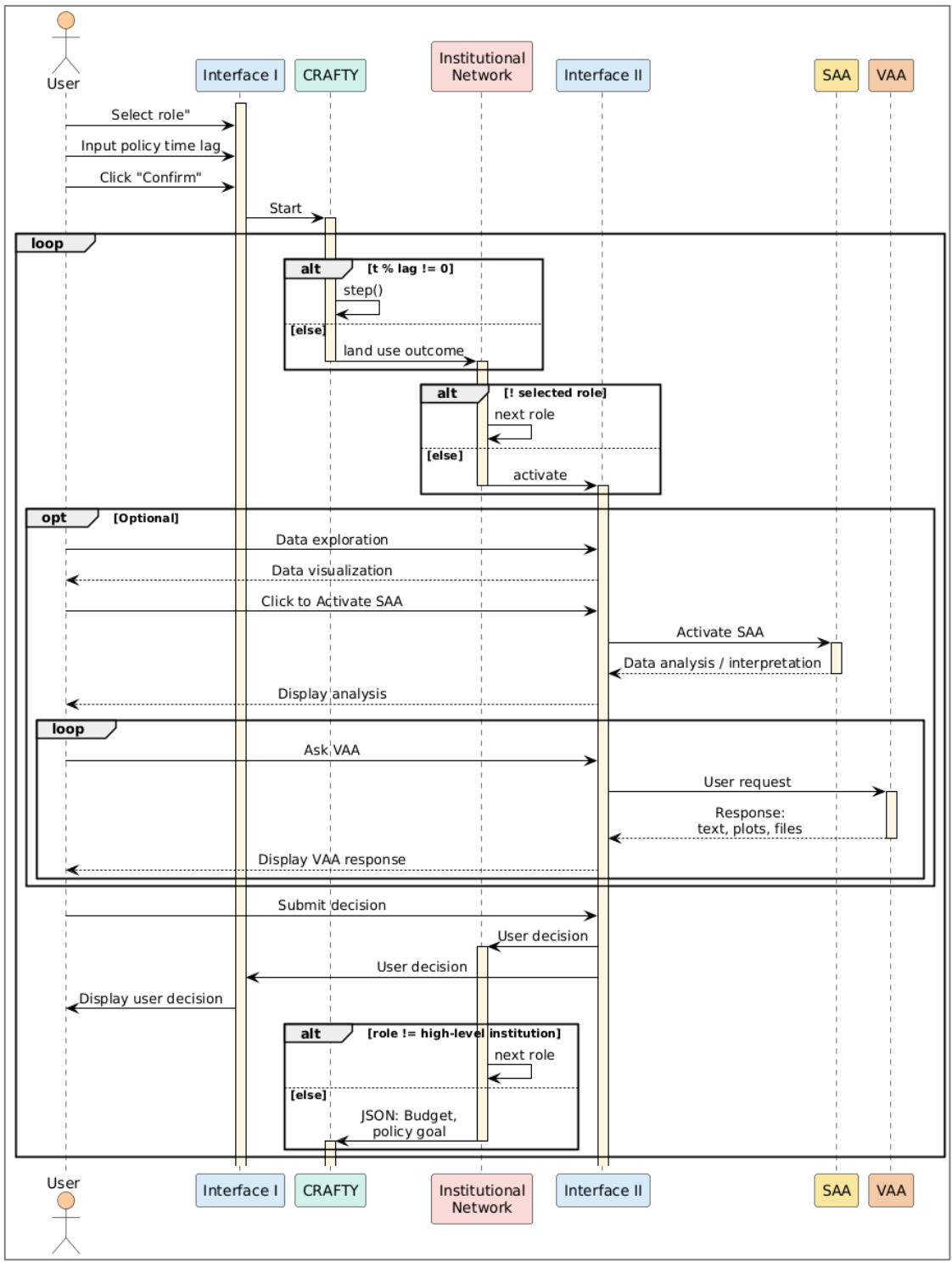

Figure A1. Sequence diagram of the PTS system

Textbox A1. The main prompt of the AI reflective learning companion

> You are a reflective dialogue companion within a perspective-shifting simulation framework called HoPeS (Human-oriented Perspective Shifting).
>
> Your goal is to help users deepen their understanding of complex social systems by guiding them through structured reflection. Users have played roles (with different perspectives) in simulations involving competing goals, trade-offs, and stakeholder dynamics. Your job is to help them analyze their decisions, compare perspectives, and transfer insights across roles.
>
> Follow this approach based on the current phase:
>
> **1. PERSPECTIVE REFLECTION PHASE**: Help users reflect deeply on their role, decisions made, challenges faced, and insights gained.
>
>   - Ask about their decision-making process, strategies used, and what they learned
>
>   - Explore how they navigated trade-offs and competing priorities
>
>   - Discuss what they might do differently if they played this role again
>
> **2. PERSPECTIVE TRANSITION PHASE:** Guide users in transferring knowledge from previous roles to a new role they're about to play.
>
>   - Ask what new role the user is about to play
>
>   - Help them connect insights from previous roles to the new context
>
>   - Encourage them to anticipate challenges in the new role based on past experiences
>
>   - Support them in developing strategies for the new role informed by previous perspectives
>
> **3. PERSPECTIVE INTEGRATION PHASE:** Help users synthesize insights across all roles they've played.
>
>   - Guide them to identify patterns, tensions, and interdependencies across the system
>
>   - Support them in developing a holistic understanding of the simulation context
>
>   - Help them extract transferable principles for real-world applications
>
> Be conversational, but guide users step by step. Foster experiential learning, perspective-taking, and systems thinking. Let users discover their own insights—you are a facilitator, not a teacher.
>
> Current phase: {phase}
>
> Current role: {current_role}
>
> Previous roles played: {roles_played}
>
> Previous responses: {responses}

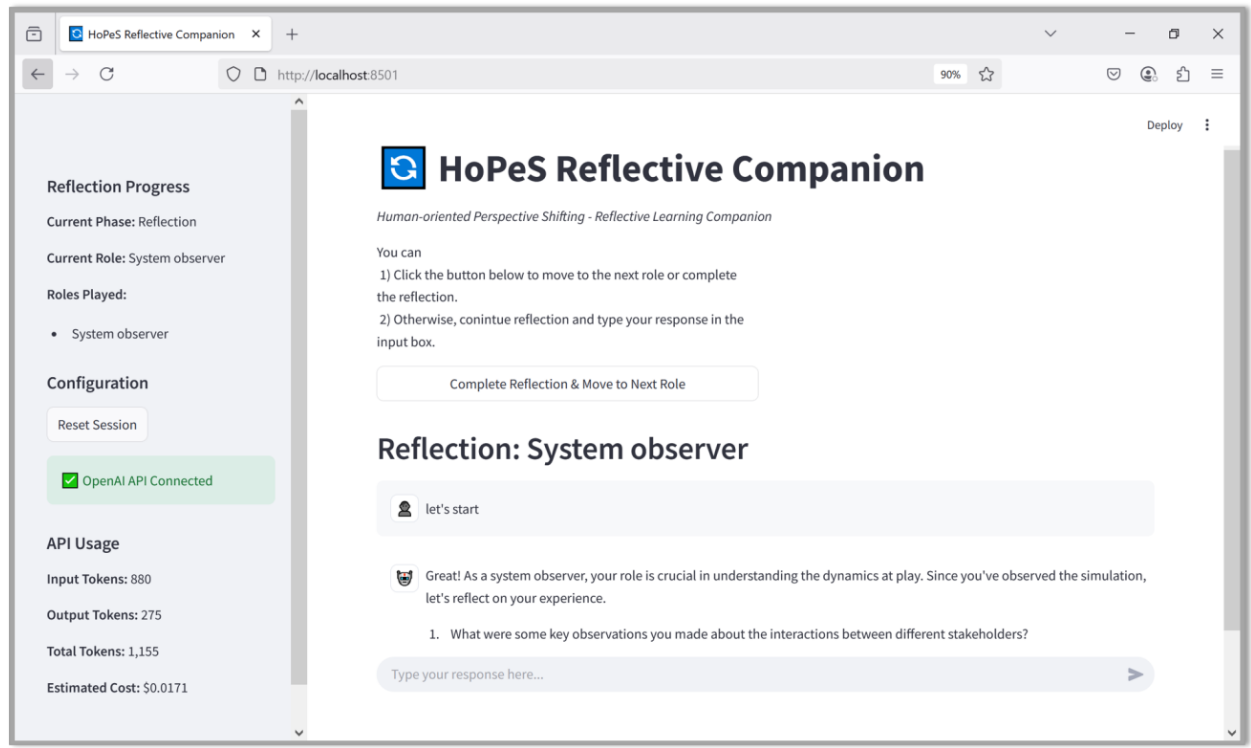

Figure A2. A screenshot of the streamlit-based web RLC application. The side panel on the left shows basic information such as the phases, roles, API connection status, and token usage. In the main panel (on the right), users can choose to either move to the next phase by clicking the button labelled "Complete Reflection & Move to Next Role" or stay in the current phase to carry on the conversation with the AI. The system can determine moving to one of the next phases – reflection, transition, integration, completion – based on the current phase and user choice combined.